\documentclass[journal]{IEEEtran}
\usepackage{amsmath,amsfonts}
\usepackage{algorithmic}
\usepackage{algorithm}
\usepackage{array}
\usepackage{textcomp}
\usepackage{stfloats}
\usepackage{url}
\usepackage{verbatim}
\usepackage{graphicx}
\usepackage{xcolor}
\usepackage{cite}
\usepackage{capt-of}
\usepackage[caption=false,font=footnotesize]{subfig}
\usepackage{tabularx}
\usepackage{booktabs}
\usepackage{multirow}
\definecolor{StableCommonFeat}{HTML}{2E7D32} 
\definecolor{ShiftedCommonFeat}{HTML}{B45F06} 
\definecolor{UniqueFeat}{HTML}{6A1B9A} 

\newcommand{\gcf}[1]{\textcolor{ShiftedCommonFeat}{#1}}
\newcommand{\uf}[1]{\textcolor{UniqueFeat}{\textit{#1}}}

\usepackage{tikz}

\newcommand{\circnum}[1]{%
  \tikz[baseline=(n.base)]{
    \node[
      circle,
      draw,
      line width=0.3pt,
      inner sep=0pt,
      minimum size=0.9em,
      font=\scriptsize
    ] (n) {#1};
  }%
}

\begin{document}

\title{RAG-HAR+: Towards Cost-Efficient LLM-Based Human Activity Recognition for Edge Deployment }

\author{
Hansi Karunarathna,
Nirhoshan Sivaroopan,
Chamara Madarasingha,
Anura Jayasumana,
and Kanchana Thilakarathna%
\thanks{Hansi Karunarathna and Nirhoshan Sivaroopan contributed equally to this work.}%
\thanks{Hansi Karunarathna, Nirhoshan Sivaroopan and Kanchana Thilakarathna are with the University of Sydney, Australia.}%
\thanks{Chamara Madarasingha is with Curtin University, Australia.}%
\thanks{Anura Jayasumana is with Colorado State University, USA.}%
}

\maketitle

\begin{abstract}
Human Activity Recognition (HAR) from wearable sensors supports applications in healthcare, rehabilitation, fitness tracking, and smart environments. Yet, existing deep learning approaches require dataset-specific training, large labeled corpora, and repeated adaptation to new sensor settings or activity taxonomies. Retrieval-Augmented Generation for Human Activity Recognition (RAG-HAR) addresses this by framing HAR as a training-free, retrieval-augmented task, in which statistical descriptions of sensor windows are used to retrieve similar labeled examples that guide LLM-based classification. We introduce RAG-HAR+, a retrieval-first and cost-optimized extension that strengthens retrieval while reducing dependence on LLM-based inference. RAG-HAR+ uses an offline Retrieval Designer Agent to design dataset-specific feature groups from a diverse pool of motion descriptors, enabling sensor windows to be compared using features better aligned with dataset-specific activity patterns. During inference, RAG-HAR+ uses majority voting over retrieved neighbors for samples with strong retrieval evidence and defers only uncertain cases to an LLM-based Ambiguity Resolver Agent. Across six HAR benchmarks, RAG-HAR+ maintains competitive or improved performance while reducing LLM usage, token consumption, and inference time. We further extend the RAG-HAR mobile prototype to demonstrate the practical feasibility of retrieval-first, LLM-assisted HAR in mobile sensing scenarios.
\end{abstract}


\section{Introduction}\label{sec:introduction}

Human Activity Recognition (HAR) from wearable and mobile sensor data underpins continuous monitoring, anomaly detection, and personalized intervention across healthcare, rehabilitation, fitness tracking, and industrial workflow monitoring~\cite{bulling2014tutorial,lara2013survey,skoda}. Because these applications run continuously on resource-constrained wearables, smartphones, and edge gateways, the practical value of a HAR system depends not only on recognition accuracy but also on per-inference computational cost, latency, and connectivity requirements.

Deep learning (DL) has become the dominant paradigm for HAR, with convolutional, recurrent, and attention-based models achieving strong benchmark performance~\cite{mccnn,deepconvlstm,metier}. However, DL-based HAR depends on dataset-specific training pipelines and hyperparameter tuning, degrades under domain shift across subjects, placements, and devices, and requires large labeled corpora that are costly to collect and annotate. As a result, HAR still lacks a solution that is simultaneously training-free, generalizable, and scalable. Large Language Models (LLMs) offer a different path for learning. Pretrained on broad corpora, they bring semantic knowledge of human activities, strong pattern matching that may support cross-domain generalization, and reasoning over context-dependent tasks. However, raw sensor streams cannot be supplied to an LLM directly because they are continuous, multi-channel time series, whereas LLMs operate over tokenized text and long windows of time series quickly exhaust the context budget. LLMs also lack grounding in activity-specific knowledge (sensor placement, activity definitions, and examples of easily confused motions) that is needed for accurate prediction. Retrieval-Augmented Generation (RAG) addresses both issues by grounding the model in concrete labeled evidence. 

Our prior work, RAG-HAR~\cite{raghar} applies this idea to HAR: it describes each sensor window with multi-segment-view statistics, indexes these descriptions in a vector database, retrieves semantically similar labeled examples, and uses the retrieved context to guide LLM-based classification, recognizing activities without training or fine-tuning a dataset-specific model. This retrieval-augmented formulation is attractive for training-free, scalable HAR, but its current realization of time-series analysis carries two limitations that become acute under continuous edge deployment. First, the retrieval representation is fixed: predefined temporal views are always described with the same basic statistics (mean, max, min, std, p25, p75, median) assuming one segment-level representation suffices across all activities and datasets. In practice, different motions call for different retrieval cues (i.e,. locomotion is better characterized by periodicity and frequency content, postures by signal magnitude and orientation, and short industrial gestures by intensity, shape, or local motion change) so a fixed temporal-view, basic-statistics descriptor may miss the most discriminative evidence. Second, the LLM is invoked for every test sample. In a typical deployment, sensing runs on a wearable or phone while the LLM is hosted and reached over the network, so each prediction incurs a network round-trip and a latency that grows with prompt length, and the retrieval-augmented prompt is already large, carrying the query description together with multiple retrieved exemplars and their summaries. The result is recurring token cost, variable latency, and a hard dependence on connectivity, paid even for windows whose retrieved neighbors already agree on a confident answer. For inference that repeats indefinitely over streaming data, this per-sample reliance on an online LLM, rather than accuracy, becomes the dominant barrier to deployment.

This paper presents RAG-HAR+, a retrieval-centric and cost-optimized framework for efficient edge HAR. Its central idea is to change \emph{when} and \emph{where} the LLM is used. Rather than acting as an online classifier invoked for every sample, the LLM is moved upstream in the pipeline into a one-time offline stage that designs a stronger retrieval representation and is retained online only as the Ambiguity Resolver Agent for the small fraction of ambiguous windows. Offline, given dataset context, candidate feature descriptions, and retrieval feedback, the offline Retrieval Designer Agent selects compact feature groups from a broad pool of statistical, temporal, spectral, and signal-shape descriptors; these groups are indexed as separate vector fields for multi-vector search, letting the retrieval representation adapt to the motion characteristics of each dataset. The benefit of the offline retrieval-design stage is reflected during online inference. By producing retrieved contexts that are more informative and better aligned with the query activity, the system can classify most windows directly through majority voting over the retrieved neighbors. The Ambiguity Resolver Agent is therefore invoked only when the retrieved evidence is ambiguous. Online computational cost therefore scales with prediction difficulty rather than with the number of test samples. The offline design cost is incurred once per dataset and amortized across all later inference, preserving the training-free nature of the framework while substantially lowering deployment cost and latency.

The main contributions of this paper are as follows:

\begin{itemize}
    
    \item We reframe LLM usage in retrieval-augmented HAR as a two-stage process: a one-time offline retrieval-refinement stage, where the LLM designs dataset-specific feature groups, and an online Ambiguity Resolver Agent, where the LLM is invoked only when retrieved neighbors do not provide a clear majority-vote prediction. Thus, online inference cost scales with prediction ambiguity rather than the number of test samples.

    \item We propose an iterative, LLM-guided Retrieval Designer Agent that uses dataset context and retrieval feedback to construct compact multi-vector retrieval groups from statistical, temporal, spectral, and signal-shape features, replacing fixed segment-level statistical groups.

    \item We implement a smartphone-based prototype that embeds RAG-HAR+ as an activity-recognition widget in a mobile fitness application, demonstrating on-device latency and the practical deployability of retrieval-first, LLM-assisted HAR in a mobile edge setting.

    \item We provide an extended evaluation across six HAR benchmarks, with ablations isolating the different component of Retrieval Designer Agent and Ambuiguity Resolver Agent. RAG-HAR+ improves performance on four datasets, remains competitive on the other two, and substantially reduces online LLM usage by 89.3--99.9\%, quantifying the accuracy--cost--latency trade-off.
\end{itemize}
\section{Related Work}\label{sec:related-work}

\subsection{Large Language Models for HAR}

LLMs have recently been explored for time-series analysis and sensor-based reasoning~\cite{llmtime,timegpt,timellm}, making them attractive for HAR, where low-level sensor patterns must be mapped to high-level activities. Recent methods connect sensor data and LLMs through different forms of adaptation: SensorLLM uses channel-specific tokens and task-aware tuning~\cite{sensorllm}, LLM4HAR combines sensor adaptation and efficiency modules~\cite{llm4har}, and PH-LLM fine-tunes an LLM on expert case studies and wearable data for personalized coaching~\cite{phllm}. Other works pair learned sensor encoders with LLM reasoning, such as SimCLR-based representations for comparing test and reference samples~\cite{simclrllm}, and extend LLMs to related sensing tasks including user recognition and activity-event description~\cite{chen2024towards,eventdescriptor}. While these studies show the promise of LLMs for sensor reasoning, they require fine-tuning, task-specific adaptation, or learned encoders, reducing the training-free advantage and limiting adaptability to new sensor configurations or activity taxonomies. Direct prompting methods such as HARGPT avoid training but degrade on fine-grained activities and higher-dimensional sensor inputs~\cite{hargpt}. These limitations motivate grounding LLM-based HAR in relevant labeled examples from the target sensing context.

\subsection{Retrieval-Augmented HAR}

RAG grounds LLM reasoning in external evidence by retrieving relevant examples or documents and providing them as context for prediction. In HAR, this allows a query sensor window to be compared with similar labeled activity windows from the target dataset, giving the LLM concrete examples of the motion patterns it must distinguish. Our prior work, RAG-HAR~\cite{raghar}, introduced a training-free retrieval-augmented framework that computes lightweight statistical descriptors from sensor windows, retrieves similar labeled samples from a vector database, and uses the retrieved context for LLM-based activity classification. However, RAG-HAR relies on fixed segment-level statistical views for retrieval and invokes the LLM for every test sample. RAG-HAR+ extends this line of work by improving retrieval and reducing online LLM usage. Instead of fixed temporal segment views and the same basic statistics across datasets, it uses LLM-guided feature-group selection to build adaptive retrieval representations from a broader feature pool. It then uses retrieval confidence to classify clear samples directly through neighbor voting and routes only ambiguous samples to the Ambiguity Resolver Agent. Thus, the LLM shifts from being a mandatory per-sample classifier to serving as an offline Retrieval Designer Agent and an online fallback for ambiguous cases.
\section{Methodology}
\label{sec:methodology}

\begin{figure*}[t]
\centering
\includegraphics[width=\textwidth]{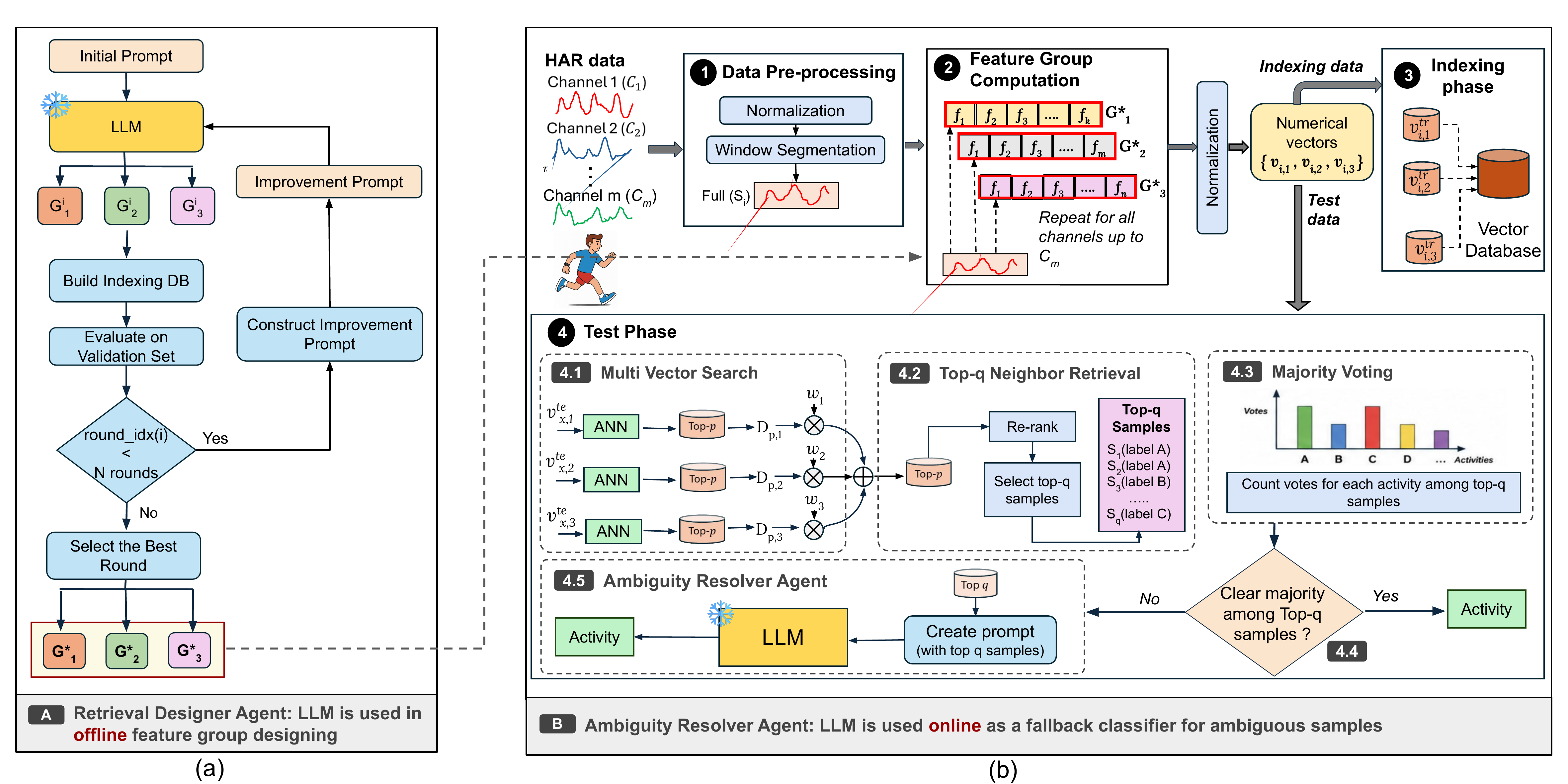}
\vspace{-6mm}
\caption{Overview of RAG-HAR+.
(a) Offline Retrieval Designer Agent for selecting dataset-specific feature groups.
(b) Feature indexing and inference pipeline, where training samples are indexed for retrieval and test samples are classified using majority voting or the Ambiguity Resolver Agent.}
\label{fig:rag_har_plus_architecture}
\vspace{-4mm}
\end{figure*}

\subsection{RAG-HAR+ Overview}
\label{sec:raghar_plus_overview}

RAG-HAR+ extends RAG-HAR by introducing an adaptive retrieval representation and a cost-optimized inference strategy. The proposed
framework is training-free. In this work, \textit{training-free} means
that RAG-HAR+ does not train or fine-tune a dataset-specific deep
learning model for activity classification. The framework still uses
labeled training samples to construct the retrieval index and a
validation split to select feature groups and retrieval weights, but
these steps configure the retrieval pipeline rather than train a
dataset-specific classifier or fine-tune the LLM.

RAG-HAR+ represents HAR windows using lightweight numerical feature vectors for retrieval and converts selected windows into structured feature descriptions only when LLM fallback is required for classification. Unlike RAG-HAR, which uses the same fixed set of statistical descriptors over predefined temporal segments and invokes the LLM for every test sample, RAG-HAR+ uses the LLM in two distinct roles. First, the \textit{Retrieval Designer Agent} uses the LLM offline to select dataset-specific feature groups for retrieval. Second, the \textit{Ambiguity Resolver Agent} uses the LLM online to resolve only the ambiguous test samples.

Fig.~\ref{fig:rag_har_plus_architecture} summarizes the overall RAG-HAR+ workflow. As shown in Fig.~\ref{fig:rag_har_plus_architecture}(a), the LLM is first used offline by the Retrieval Designer Agent to select dataset-specific feature groups. Starting from an initial prompt, the agent proposes three candidate feature groups, which are used to build an indexing database and are evaluated on a validation set. The validation feedback is then used to construct an improvement prompt for the next round. After multiple rounds, the best-performing feature groups, denoted as $G_1^{*}$, $G_2^{*}$, and $G_3^{*}$, are selected as the final feature groups.

Fig.~\ref{fig:rag_har_plus_architecture}(b) shows how the selected feature groups are used during indexing and inference. The pipeline begins with data preprocessing~\circnum{1}, where raw HAR sensor streams are normalized and segmented into fixed-size windows. The selected feature groups from Fig.~\ref{fig:rag_har_plus_architecture}(a) are then computed for each window across all sensor channels~\circnum{2}. The resulting feature-group vectors are normalized and stored in the vector database together with their activity labels~\circnum{3}. During the test phase~\circnum{4}, the same preprocessing and feature-group computation steps are applied to the query window. RAG-HAR+ then performs multi-vector search, re-ranks the retrieved samples, selects the Top-$q$ neighbors, and applies majority voting over their labels, corresponding to steps~\circnum{4.1}--\circnum{4.3}. If a single activity label obtains the highest vote count, the activity is predicted directly from retrieval~\circnum{4.4}. If two or more labels share the highest vote count, the sample is treated as ambiguous and routed to the Ambiguity Resolver Agent~\circnum{4.5}. This selective fallback mechanism reduces unnecessary online LLM calls while retaining LLM reasoning for difficult cases.

\subsection{Offline Feature-Group Design}
\subsubsection{\textbf{Candidate Feature Pool Generation}}
\label{sec:feature_pool}

RAG-HAR+ defines a candidate feature pool as the search space for constructing feature groups. The pool contains feature descriptors spanning statistical, temporal, spectral, motion-intensity, and signal-shape categories \cite{Barandas2020}. Statistical features capture distributional properties of each channel, temporal features describe local signal changes and movement dynamics, spectral features capture frequency-domain behavior, and shape-based features describe waveform morphology. The pool is provided to the Retrieval Designer Agent as a flat list of feature names and short descriptions, from which the agent selects the feature groups.

\subsubsection{\textbf{Retrieval Designer Agent}}
\label{subsec:retriver-designer-agent}
RAG-HAR+ uses the LLM offline as a Retrieval Designer Agent to construct compact, dataset-specific feature groups. This is a one-time process performed on each dataset before test-time inference. Given the candidate feature pool, the agent proposes three feature groups at iteration $i$, denoted as $\mathcal{G}^{(i)}=\{G_1^{(i)},G_2^{(i)},G_3^{(i)}\}$. Each group contains a subset of features from the pool and forms one vector field in the multi-vector retrieval index. The choice of three groups is motivated by the ablation results in Section~\ref{sec:ablation_number_of_feature_views}, where three groups achieve the best retrieval performance.

The selected groups are not restricted to a single feature category. Instead, each group may combine statistical, temporal, spectral, motion-intensity, and signal-shape descriptors to provide complementary representations of the same sensor window. This allows retrieval to compare activity windows from multiple perspectives rather than relying on a single fixed descriptor set. Each selected feature is assigned to only one group to reduce redundancy across feature groups and keep the retrieval representation compact.

\begin{figure*}[t]
    \centering
    \includegraphics[width=1\linewidth]{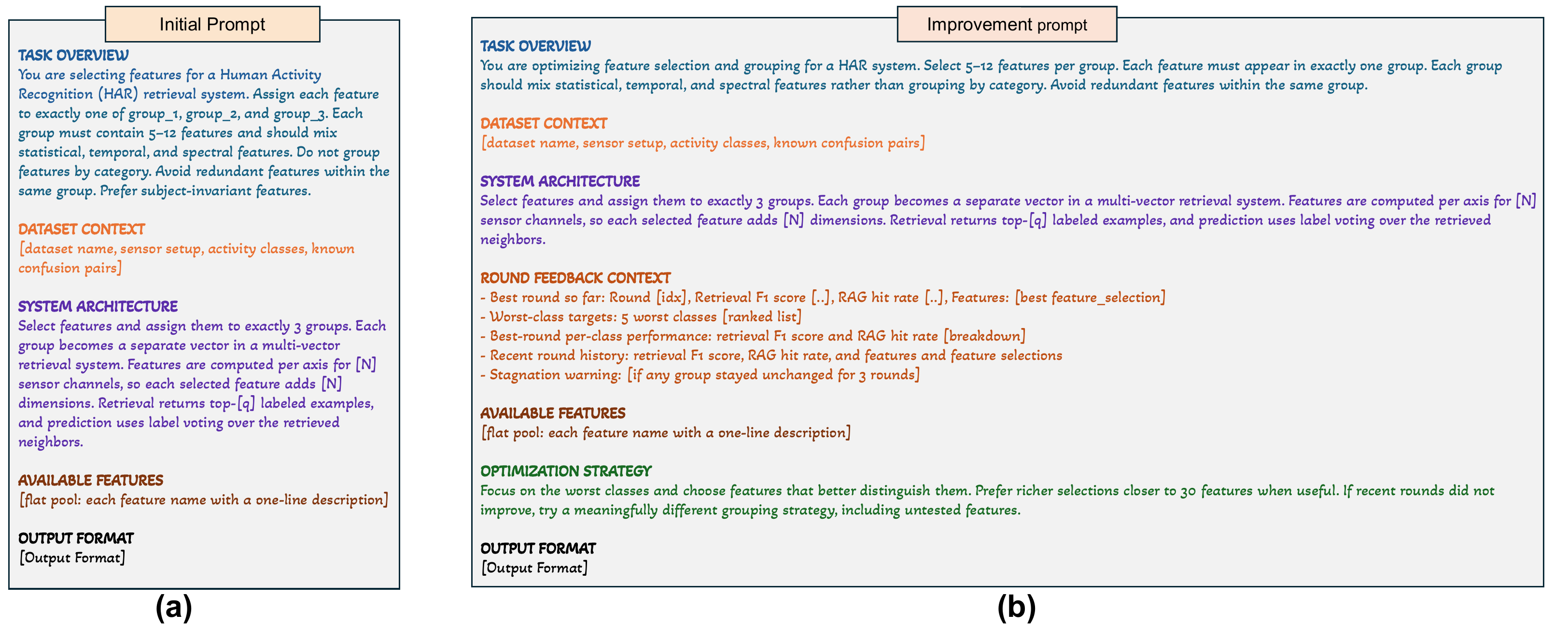}
    \vspace{-6mm}
    \caption{Prompt structure used by the Retrieval Designer Agent. 
    (a) Initial prompt used to generate the first feature-group configuration. 
    (b) Improvement prompt used in subsequent rounds to refine feature groups based on retrieval feedback.}
    \label{fig:feature_group_prompts}
    \vspace{-6mm}
\end{figure*}

As shown in Fig.~\ref{fig:rag_har_plus_architecture}(a), the feature-group design process is performed iteratively using an initial prompt followed by improvement prompts. The structure of these prompts is illustrated in Fig.~\ref{fig:feature_group_prompts}. The initial prompt in Fig.~\ref{fig:feature_group_prompts}(a) provides the Retrieval Designer Agent with the task overview, dataset context, system constraints, candidate feature pool, and expected output format. The dataset context includes information such as the dataset name, sensor setup, sensor locations, activity classes, and known confusing activity pairs. The system constraints specify that the agent should assign features to three compact groups, avoid assigning the same feature to multiple groups, and reduce redundant or subject-dependent features.

At iteration $i$, the proposed configuration $\mathcal{G}^{(i)}$ is used to build a temporary multi-vector retrieval index on the training data and is evaluated on a validation split. The validation stage uses two metrics: retrieval F1 score and RAG hit rate. The retrieval F1 score measures the classification performance obtained by applying majority voting over the retrieved neighbors and comparing the predicted activity label with the ground-truth label. RAG hit rate measures whether the correct activity label appears among the retrieved candidates. Thus, retrieval F1 score evaluates whether retrieval alone is sufficient for direct classification, while RAG hit rate evaluates whether the retrieved neighbors contain at least one correct-label example that can support the Ambiguity Resolver Agent when fallback is needed.

After each round, an improvement prompt is constructed using the validation feedback,  as illustrated in Fig.~\ref{fig:feature_group_prompts}(b). This prompt includes the best-performing configuration so far, overall retrieval F1 score, RAG hit rate, weak activity classes, per-class retrieval behavior, recent feature-group selection history, and stagnation warnings when the selections do not improve across rounds. It also instructs the agent to focus on poorly performing classes, choose features that better distinguish confusing activities, avoid redundant descriptors, and explore alternative grouping strategies when needed.  This allows the LLM to adjust the feature groups based on observed retrieval errors, for example by adding features that better distinguish confusing activities or removing redundant features that do not improve retrieval. After multiple refinement rounds, the configuration with the strongest validation Retrieval F1 score is selected as the final retrieval representation, denoted as $\mathcal{G}^{*}=\{G_1^{*}, G_2^{*}, G_3^{*}\}$.

Table~\ref{tab:selected_features_detail} summarizes the final feature groups selected by the Retrieval Designer Agent for all six datasets. The table shows that the agent does not simply reuse a fixed descriptor set across datasets. Instead, it consistently preserves some generally useful descriptors, such as standard deviation, interquartile range, and autocorrelation, while adapting the remaining descriptors and their group assignments to the sensing setup and activity classes of each dataset. This supports the intended role of the agent: to design compact, non-overlapping, and dataset-specific feature groups that provide complementary retrieval representations.

\subsection{Indexing and Inference Pipeline}
\subsubsection{\textbf{Data Preprocessing}}
\label{sec:data_preprocessing}

Given raw HAR sensor streams, RAG-HAR+ first applies dataset-specific preprocessing to generate fixed-size windows. Each dataset contains multiple sensor channels, such as accelerometer or gyroscope axes, collected from one or more body locations or devices. Let $C=\{C_1,C_2,\ldots,C_m\}$ denote the set of sensor channels. As shown in Fig.~\ref{fig:rag_har_plus_architecture}(b), Step~\circnum{1}, data preprocessing consists of two main steps. \textit{i})~\textbf{Normalization:} All raw sensor channels are standardized using Z-score normalization. For each channel, the mean and standard deviation are estimated exclusively from the training split and applied unchanged to the test splits. \textit{ii})~\textbf{Sliding window-based segmentation:} Following the protocols used in prior HAR studies \cite{khaertdinov2021deep, suh2022adversarial, ahmad2023alae, haresamudram2023investigating}, the normalized time-series data are partitioned into fixed-size windows using a sliding window with a predefined window size and step size. We denote the resulting labeled window set as
$\mathcal{D}=\{(S_i,y_i)\}_{i=1}^{N}$, where $S_i$ denotes the $i$-th sensor window and $y_i$ denotes its activity label. This process captures short-term temporal context and produces fixed-size units for feature computation in Step~\circnum{2} of Fig.~\ref{fig:rag_har_plus_architecture}(b).

\subsubsection{\textbf{Feature-Group Computation and Vector Database Indexing}}

\label{sec:indexing}

After the final feature-group configuration is selected, RAG-HAR+ computes feature-group vectors for the labeled training windows and stores them in the vector database. Let the labeled indexing set be denoted as
$\mathcal{D}^{\mathrm{tr}}=\{(S_i,y_i)\}_{i=1}^{N_{\mathrm{tr}}}$,
where $S_i$ is the $i$-th training window, $y_i$ is its activity label, and $i$ indexes the training windows. Let the final selected feature-group configuration be
$\mathcal{G}^{*}=\{G_1^{*},G_2^{*},G_3^{*}\}$,
where $k\in\{1,2,3\}$ indexes the selected feature groups.

For each training window $S_i$, RAG-HAR+ computes one feature vector from each selected group $G_k^{*}$. The features in $G_k^{*}$ are computed across all sensor channels. If $G_k^{*}$ contains $d_k$ features and the dataset has $m$ sensor channels, the resulting vector for group $k$ has $d_k \times m$ dimensions. Each feature-group vector is then normalized using the per-dimension mean and standard deviation computed from the indexing split.

Thus, each indexed window $S_i$ is represented by three normalized feature-group vectors:
\begin{equation}
\mathbf{V}_i^{\mathrm{tr}} =
\left\{
\mathbf{v}_{i,1}^{\mathrm{tr}},
\mathbf{v}_{i,2}^{\mathrm{tr}},
\mathbf{v}_{i,3}^{\mathrm{tr}}
\right\},
\end{equation}
where $\mathbf{v}_{i,k}^{\mathrm{tr}}$ denotes the normalized vector generated from window $S_i$ using feature group $G_k^{*}$.

RAG-HAR+ then builds a multi-vector index using these feature-group vectors. Here, indexing refers to storing each labeled training window as a retrievable reference item in the vector database. Unlike a single-vector index, RAG-HAR+ stores each window as one database record with three separate vector fields, one for each selected feature group:
\begin{equation}
r_i =
\left(
\mathbf{v}_{i,1}^{\mathrm{tr}},
\mathbf{v}_{i,2}^{\mathrm{tr}},
\mathbf{v}_{i,3}^{\mathrm{tr}},
y_i
\right),
\end{equation}
where $r_i$ is the indexed database record for window $S_i$. Therefore, each activity window has a single label but can be retrieved through three complementary feature groups.

During inference, the same feature groups and normalization statistics are applied to the query window to generate three query vectors. Each query vector is searched against the corresponding vector field in the database, and the retrieved results are combined during weighted multi-vector retrieval.

\subsubsection{\textbf{Weighted Multi-Vector Search}}

\label{sec:multi_vector_search}

For an unseen test window $S_x$, RAG-HAR+ applies the same preprocessing, feature-group computation, and normalization steps used during indexing. This produces three query vectors: 
\begin{equation}
    \mathbf{V}_x=\{\mathbf{v}_{x,1}^{\mathrm{te}},
    \mathbf{v}_{x,2}^{\mathrm{te}},
    \mathbf{v}_{x,3}^{\mathrm{te}}\}.
\end{equation}

Approximate nearest-neighbor search is performed separately for each feature-group vector. For each selected feature group $G_k^{*}$, the corresponding query vector $\mathbf{v}_{x,k}^{\mathrm{te}}$ is used to retrieve the Top-$p$ candidate windows from the indexed vectors of the same feature group. We denote the candidate set retrieved from feature group $k$ as $D_{x,k}^{(p)}$. The candidates from the three feature groups are then merged and re-ranked to select the final Top-$q$ neighbors. Let $s_{i,k}$ denote the similarity between the query window $S_x$ and an indexed window $S_i$ under feature group $k$.

\begin{equation}
    \rho_i = \sum_{k=1}^{3} w_k^{*} s_{i,k},
\end{equation}
where $w_k^{*}$ is the optimized weight assigned to feature group $k$, $w_k^{*} \geq 0$, and $\sum_{k=1}^{3}w_k^{*}=1$.

The weights $\{w_1^{*},w_2^{*},w_3^{*}\}$ are selected through a validation-set grid search over candidate weight combinations as described in Section~\ref{sec:ablation_feature_group_weights}. The validation set is constructed only from the training split, and the weight configuration that gives the best validation retrieval objective is fixed for final test-set inference. The merged candidates are re-ranked using $\rho_i$, and the Top-$q$ samples are selected as the final retrieved neighbor set:
\begin{equation}
    D_x^{(q)} = \{(S_j,y_j,\rho_j)\}_{j=1}^{q}.
\end{equation}

\subsubsection{\textbf{Retrieval-Based Classification}}

\label{sec:classification}

After Top-$q$ retrieval, RAG-HAR+ decides whether the retrieved evidence alone is sufficient to classify the query or whether the query should be deferred to the Ambiguity Resolver Agent. This decision is based on the agreement among the labels of the retrieved neighbors. Note that in $D_x^{(q)}$ the score $\rho_j$ is used only to select and order the Top-$q$ neighbors during retrieval; in the voting stage, it plays no further role, and each retained neighbor contributes a single, equally weighted vote.
For each activity class $c \in \mathcal{Y}$, the vote count over the retrieved neighbor set is computed using an indicator function. Here, $\mathbb{I}(y_j=c)$ is defined as:
\begin{equation}
\mathbb{I}(y_j=c)=
\begin{cases}
1, & \text{if } y_j=c,\\
0, & \text{otherwise}.
\end{cases}
\end{equation}

The vote count for class $c$ is then:
\begin{equation}
V(c)=
\sum_{(S_j,y_j,\rho_j)\in D_x^{(q)}}
\mathbb{I}(y_j=c).
\end{equation}

The retrieval-based prediction is:
\begin{equation}
\hat{y}_{\mathrm{ret}}=
\arg\max_{c\in\mathcal{Y}} V(c).
\end{equation}

RAG-HAR+ then applies a deterministic routing rule. If one activity label obtains a unique highest vote count, the retrieved neighbors are in clear agreement and the query is classified directly from retrieval, $\hat{y}= \hat{y}_{\mathrm{ret}}$. If two or more labels share the highest vote count,
majority voting is undecided, and the query is routed to the Ambiguity Resolver Agent.

We deliberately adopt parameter-free majority voting as the routing rule. 
Because the database contains many close samples, we empirically observed that the similarity scores among the top retrieved neighbors often do not have a large margin. Majority voting avoids additional hyperparameters, requires no calibration data, and keeps the routing rule deterministic and training-free.

\subsubsection{\textbf{Ambiguity Resolver Agent}}

For queries routed to the Ambiguity Resolver Agent, RAG-HAR+ constructs a structured prompt using the query window, the retrieved examples, their activity labels, and the dataset-specific activity definitions. Although adaptive feature groups are used for retrieval, these high-dimensional vectors are not directly exposed to the LLM. This is because LLMs reason more effectively over structured textual descriptions with semantically meaningful fields than over dense numerical vectors optimized for similarity search.

Instead, we use the segment-wise statistical prompt format from RAG-HAR, where each window is described using basic descriptors computed over the full window and the Start, Mid, and End sub-segments. This introduces only a small additional computational cost because these descriptors are lightweight and are computed only for the query and the Top-$q$ retrieved examples when fallback is triggered. The separation allows RAG-HAR+ to use adaptive feature groups for retrieving stronger labeled neighbors, while presenting the Ambiguity Resolver Agent with a compact and temporally ordered representation suitable for comparison and reasoning.

The LLM then selects the most likely activity label from the valid label set, $\hat{y}=\hat{y}_{\mathrm{LLM}}$. This selective fallback mechanism preserves LLM reasoning for ambiguous cases while avoiding unnecessary LLM calls when the retrieved neighbors already provide sufficient evidence.

\begin{table*}[t]
\centering
\caption{Selected feature groups for each dataset by the Retrieval Designer Agent. 
Orange features are common to all six datasets but appear in different feature groups across datasets. Purple italicized features are not shared by all six datasets.}
\label{tab:selected_features_detail}
\scriptsize
\setlength{\tabcolsep}{3pt}
\renewcommand{\arraystretch}{1.12}
\begin{tabularx}{\textwidth}{p{0.09\textwidth}XXX}
\toprule
\textbf{Dataset} & \textbf{$G_1^{*}$} & \textbf{$G_2^{*}$} & \textbf{$G_3^{*}$} \\
\midrule

USC-HAD
& \gcf{mid-frequency bandpower}, \uf{high-frequency bandpower}, \gcf{dominant frequency}, \gcf{dominant-frequency ratio}, \uf{autocorrelation at lag 10}, \uf{peak count}, \uf{peak-frequency ratio}, \uf{skewness}, \uf{positive ratio}, \uf{mean second difference}, \uf{spectral centroid}
& \gcf{sample entropy}, \uf{count above mean}, \uf{standard deviation}, \gcf{mean absolute change}, \uf{very-low-frequency bandpower}, \uf{low-frequency bandpower}, \uf{low--high bandpower ratio}, \uf{spectral flatness}, \uf{mean-crossing rate}, \uf{spectral entropy}
& \uf{RMS}, \uf{total power}, \uf{crest factor}, \uf{impulse factor}, \uf{waveform length}, \gcf{turning-points ratio}, \uf{autocorrelation at lag 1}, \uf{autocorrelation at lag 5}, \uf{autocorrelation at lag 2}, \uf{interquartile range} \\

\midrule

Skoda
& \uf{standard deviation}, \uf{signal energy}, \uf{ratio beyond one standard deviation}, \uf{count above mean}, \uf{interquartile range}, \gcf{mean absolute change}, \uf{crest factor}, \uf{autocorrelation at lag 1}, \uf{zero-crossing rate}
& \uf{skewness}, \uf{kurtosis}, \gcf{sample entropy}, \uf{Hjorth complexity}, \uf{waveform length}, \gcf{turning-points ratio}, \uf{mean-crossing rate}, \uf{coefficient of variation}, \uf{spectral entropy}
& \uf{autocorrelation at lag 2}, \gcf{dominant frequency}, \gcf{dominant-frequency ratio}, \uf{low-frequency bandpower}, \gcf{mid-frequency bandpower}, \uf{spectral centroid}, \uf{spectral spread}, \uf{linear trend slope}, \uf{start--end difference} \\

\midrule

GOTOV
& \uf{standard deviation}, \uf{interquartile range}, \uf{coefficient of variation}, \uf{kurtosis}, \uf{ratio beyond one standard deviation}, \gcf{sample entropy}, \uf{spectral flatness}, \uf{spectral entropy}, \uf{very-low-frequency bandpower}, \uf{spectral slope}
& \uf{skewness}, \uf{mean change}, \uf{positive ratio}, \uf{linear trend slope}, \uf{start--end difference}, \uf{autocorrelation at lag 1}, \gcf{mean absolute change}, \uf{mean second difference}, \gcf{turning-points ratio}, \uf{waveform length}
& \gcf{dominant frequency}, \gcf{dominant-frequency ratio}, \uf{peak-frequency ratio}, \uf{autocorrelation at lag 5}, \uf{autocorrelation at lag 10}, \uf{zero-crossing rate}, \uf{mean-crossing rate}, \uf{low-frequency bandpower}, \gcf{mid-frequency bandpower}, \uf{high-frequency bandpower} \\

\midrule

MHEALTH
& \uf{standard deviation}, \uf{RMS}, \gcf{dominant frequency}, \gcf{dominant-frequency ratio}, \uf{peak-frequency ratio}, \gcf{mid-frequency bandpower}, \uf{high-frequency bandpower}, \uf{low-frequency bandpower}, \uf{low--high bandpower ratio}, \uf{total power}, \uf{autocorrelation at lag 10}, \uf{peak count}
& \uf{ratio beyond one standard deviation}, \gcf{sample entropy}, \uf{spectral flatness}, \uf{mean second difference}, \uf{count above mean}, \uf{positive ratio}, \uf{median}, \uf{interquartile mean}, \uf{spectral centroid}, \uf{spectral spread}, \uf{mean-crossing rate}, \uf{very-low-frequency bandpower}
& \gcf{mean absolute change}, \uf{Hjorth mobility}, \uf{skewness}, \uf{kurtosis}, \uf{linear trend slope}, \uf{start--end difference}, \uf{impulse factor}, \uf{clearance factor}, \gcf{turning-points ratio}, \uf{waveform length}, \uf{crest factor} \\

\midrule

HHAR
& \gcf{dominant frequency}, \gcf{dominant-frequency ratio}, \uf{peak-frequency ratio}, \uf{autocorrelation at lag 1}, \uf{zero-crossing rate}, \gcf{mid-frequency bandpower}, \uf{low--high bandpower ratio}
& \uf{linear trend slope}, \uf{start--end difference}, \uf{first-half--second-half difference}, \uf{mean second difference}, \uf{mean change}, \uf{crest factor}, \uf{impulse factor}, \uf{skewness}
& \uf{ratio beyond one standard deviation}, \gcf{sample entropy}, \uf{count above mean}, \uf{spectral flatness}, \gcf{mean absolute change}, \gcf{turning-points ratio}, \uf{interquartile mean}, \uf{trimmed mean (10\%)} \\

\midrule

PAMAP2
& \uf{signal energy}, \gcf{mean absolute change}, \uf{autocorrelation at lag 1}, \gcf{mid-frequency bandpower}, \uf{high-frequency bandpower}, \uf{low-frequency bandpower}, \uf{low--high bandpower ratio}, \gcf{dominant frequency}, \gcf{dominant-frequency ratio}, \uf{mean-crossing rate}
& \uf{ratio beyond one standard deviation}, \gcf{sample entropy}, \uf{spectral flatness}, \uf{mean second difference}, \uf{spectral entropy}, \uf{zero-crossing rate}, \uf{standard deviation}, \uf{interquartile range}, \uf{crest factor}, \uf{impulse factor}
& \uf{linear trend slope}, \uf{start--end difference}, \uf{first-half--second-half difference}, \uf{autocorrelation at lag 10}, \uf{peak count}, \uf{peak-frequency ratio}, \uf{waveform length}, \gcf{turning-points ratio}, \uf{second dominant frequency}, \uf{spectral centroid} \\

\bottomrule
\end{tabularx}
\vspace{-6mm}
\end{table*}
\section{Evaluation}
\label{sec:results}

We evaluate RAG-HAR+ on six public HAR benchmark datasets used in the original RAG-HAR study: USC-HAD~\cite{uschad}, PAMAP2~\cite{pamap2}, MHEALTH~\cite{mhealth}, GOTOV~\cite{paraschiakos2020activity}, HHAR~\cite{hhar}, and Skoda~\cite{skoda}. These datasets cover diverse sensing conditions, including single-IMU setups, multi-location body-worn sensors, smartwatch-based sensing, and high-dimensional industrial sensor arrays. They also include a wide range of activity categories, from locomotion and postural activities to daily living activities and assembly-line gestures. We follow the same preprocessing, windowing, indexing, and testing protocols used in RAG-HAR\cite{raghar} to ensure a fair comparison with the original method. \footnote{Complete details of the dataset, protocols, LLM model, database, features and hyper-parameters can be found in the Supplementary material.}

To analyze retrieval quality, we report the retrieval F1 score and RAG hit rate. Retrieval F1 score (RF1) measures the classification performance obtained by applying majority voting over the retrieved neighbors and comparing the predicted activity label with the ground-truth label. RAG hit rate (RHR) measures whether the ground-truth label appears among the Top-$q$ retrieved neighbors.

\subsection{Benchmarking RAG-HAR+}
\label{sec:benchmarking_raghar_plus}

\begin{figure*}[t]
\centering

\begin{minipage}[t]{0.45\textwidth}
\vspace{0pt}
\centering
\captionof{table}{End-to-end benchmark comparison with the SOTA non-LLM and LLM HAR baselines.}
\label{tab:benchmark_comparison}
\scriptsize
\setlength{\tabcolsep}{3pt}
\resizebox{\linewidth}{!}{%
\begin{tabular}{llcc}
\toprule
\textbf{Dataset} & \textbf{Method} &
\textbf{Acc.} &
\textbf{F1} \\
\midrule

\multirow{6}{*}{USC-HAD}
& Triplet LSTM (HTL-SB)~\cite{khaertdinov2021deep} & -- & 62.80  \\
& Transformer-like architecture ~\cite{khaertdinov2021deep} & -- & 55.00  \\
& DeepConvLSTM ~\cite{khaertdinov2021deep}    & --   & 46.00      \\
& Sensor-LLM~\cite{sensorllm}  & 62.60  & 61.20 \\
& RAG-HAR & 57.20 & 58.63 \\
& RAG-HAR+ & \textbf{60.62} & \textbf{60.12} \\

\midrule

\multirow{7}{*}{PAMAP2}
& Triplet LSTM (OTL)~\cite{khaertdinov2021deep} & -- & 90.40  \\
& b-LSTM-S ~\cite{khaertdinov2021deep}             & --                          & 86.80          \\
& ADFE ~\cite{suh2022adversarial}            & 85.69                         & 77.84      \\
& HARGPT~\cite{hargpt}       &  32.11    & 31.57  \\
& LLM as Virtual Annotators ~\cite{hota2025evaluating}& 56.70    & 53.80  \\
& RAG-HAR & \textbf{91.60} & \textbf{91.12}  \\
& RAG-HAR+ & 90.35 & 90.60 \\

\midrule

\multirow{5}{*}{MHEALTH}
& ADFE~\cite{suh2022adversarial} & 96.72 & 96.47 \\
& METIER  ~\cite{suh2022adversarial}      & 94.42              & 94.09  \\
& MC-CNN  ~\cite{suh2022adversarial}           & 89.69               & 87.42  \\
& RAG-HAR & 96.91 & 96.74  \\
& RAG-HAR+ & \textbf{98.35} & \textbf{98.19}  \\

\midrule

\multirow{5}{*}{GOTOV}
& ALAE-TAE-CutMix+~\cite{ahmad2023alae} & -- & 79.40 \\
& Attend and Discriminate ~\cite{ahmad2023alae}     & -                          & 76.20             \\
& Att. Model ~\cite{ahmad2023alae}             & -                          & 70.70       \\
& RAG-HAR & \textbf{81.55} & \textbf{79.92} \\
& RAG-HAR+ & 76.51 & 75.97 \\

\midrule

\multirow{5}{*}{HHAR}
& Enhanced CPC~\cite{haresamudram2023investigating} & -- & 59.25  \\
& CPC ~\cite{haresamudram2023investigating}   & -                          & 59.17 \\
& SimCLR ~\cite{haresamudram2023investigating}   & -                & 52.84            \\
& RAG-HAR & 58.61 & 59.86\\
& RAG-HAR+ & \textbf{62.70} & \textbf{61.02}  \\

\midrule

\multirow{5}{*}{Skoda}
& ALAE-TAE-CutMix+~\cite{ahmad2023alae} & -- & 94.80  \\
& Attend and Discriminate  ~\cite{ahmad2023alae}       & -                          & 92.80              \\
& b-LSTM-S  ~\cite{ahmad2023alae}      & -                          & 92.10             \\
& RAG-HAR & 96.04 & 95.21 \\
& RAG-HAR+ & \textbf{97.73} & \textbf{97.74} \\

\bottomrule
\end{tabular}%
\vspace{-10mm}
}
\end{minipage}%
\hfill
\begin{minipage}[t]{0.51\textwidth}
\vspace{0pt}
\centering

\includegraphics[width=\linewidth]{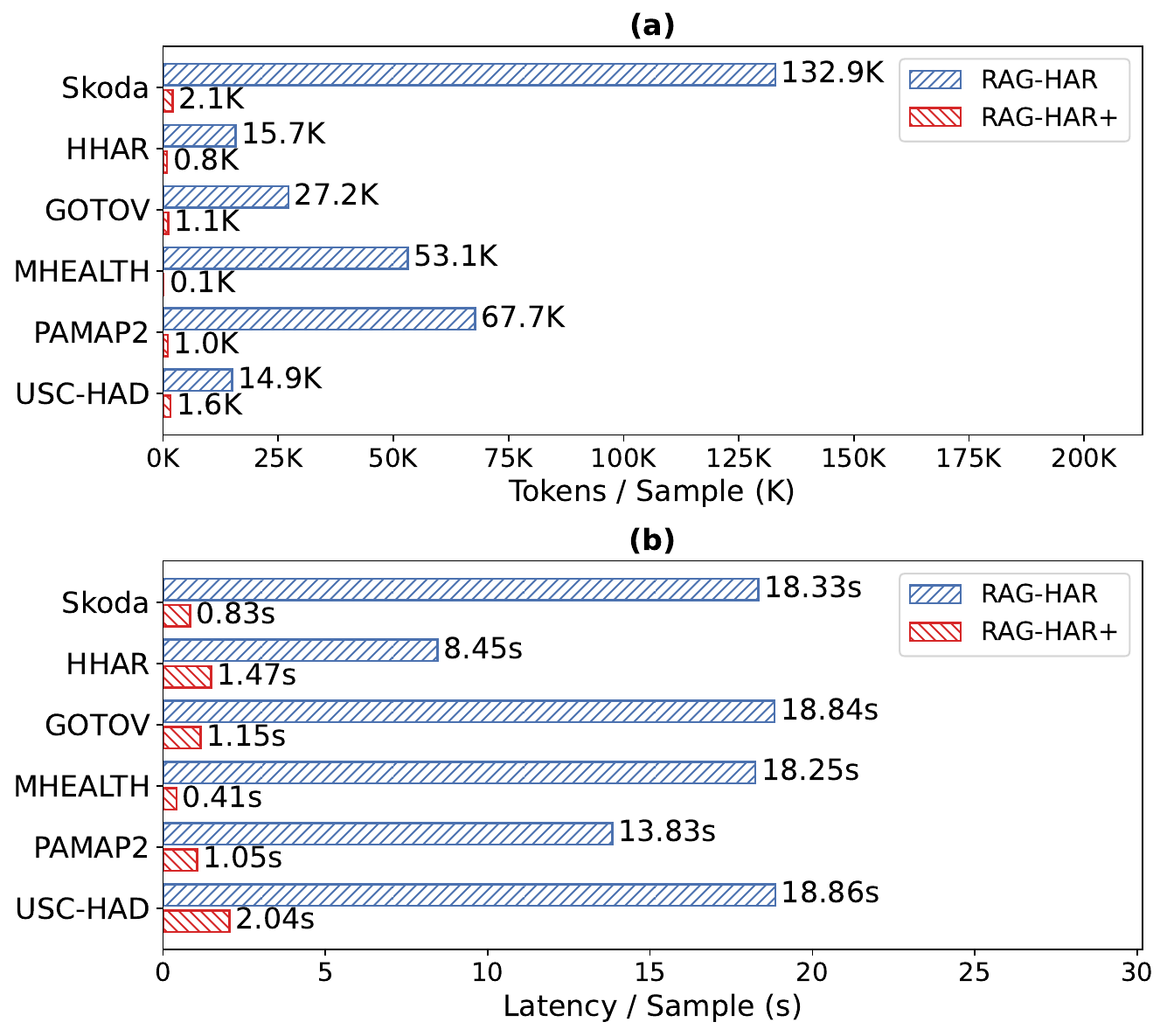}
\vspace{-8mm}
\captionof{figure}{
(a) Average online LLM token usage.
(b) Average inference latency}
\label{fig:per-sample-cost}

\vspace{1mm}

\includegraphics[width=\linewidth]{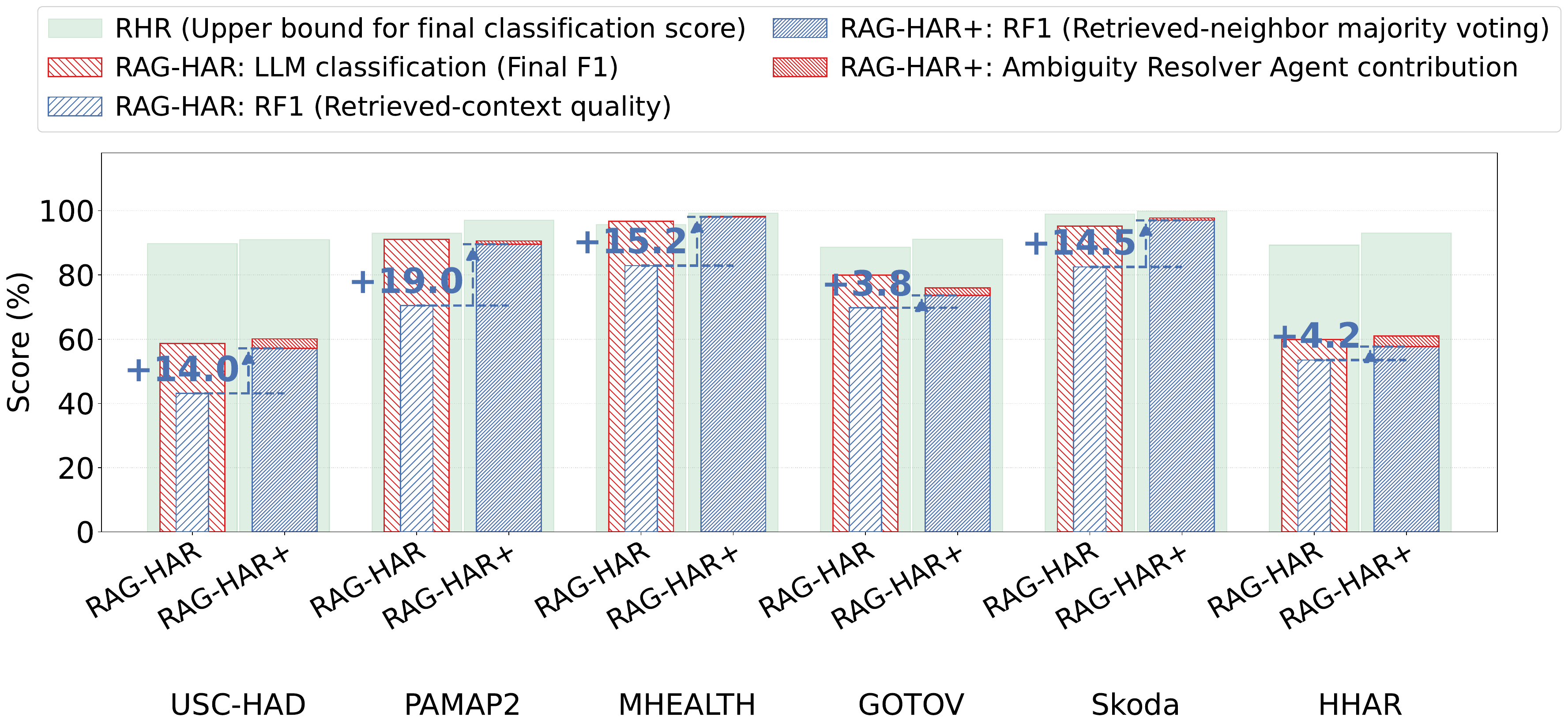}
\vspace{-8mm}
\caption{Comparison of RAG-HAR and RAG-HAR+. For RAG-HAR, the blue bars indicate retrieval F1, while the red bars show the final F1 after LLM classification. For RAG-HAR+, the blue bars show the F1 achieved through retrieved-neighbor majority voting, and the stacked red segments indicate the additional gain contributed by the Ambiguity Resolver Agent.}

\label{fig:accuracy_contribution}
\vspace{-6mm}
\end{minipage}
\vspace{-6mm}
\end{figure*}

Table~\ref{tab:benchmark_comparison} compares the RAG-HAR+ pipeline with the RAG-HAR and the strongest prior HAR baselines reported for each dataset. We mention only the best non-LLM and LLM baselines in the table. Additional baselines for RAG-HAR benchmarking are reported in \cite{raghar}. Overall, RAG-HAR+ achieves competitive performance across all six datasets while reducing the reliance on LLM-based inference. RAG-HAR+ achieves higher performance than RAG-HAR on four datasets: USC-HAD, MHEALTH, HHAR, and Skoda, demonstrating that the adaptive feature-group representation improves retrieval quality and contributes to stronger classification outcomes. Compared with prior deep learning baselines, the results further show that RAG-HAR+ remains highly competitive while preserving the training-free retrieval-augmented design of RAG-HAR.

Although the performance gains over RAG-HAR are modest on some datasets, they are noteworthy because RAG-HAR+ achieves these results with substantially lower online LLM inference. As shown in Fig.~\ref{fig:per-sample-cost}, RAG-HAR+ substantially reduces per-sample token usage and inference latency compared with RAG-HAR. Unlike RAG-HAR, which invokes the LLM for every sample, RAG-HAR+ first relies on retrieval-based classification and uses the LLM only for ambiguous cases. RAG-HAR+ therefore provides a more efficient operating point by maintaining competitive recognition performance while reducing token usage, API cost, and inference latency. This makes the proposed pipeline more suitable for resource-constrained mobile environments where continuous sensor streams can otherwise lead to repeated and costly LLM invocations.

\subsection{LLM Cost Analysis}
\label{sec:results_cost}

This subsection analyzes the LLM usage and cost of RAG-HAR+. We report offline and online costs separately because they occur at different stages of the pipeline. The offline cost corresponds to the Retrieval Designer Agent, which is executed once for each dataset configuration. After this step, the selected feature groups and weights are fixed and reused during inference. Therefore, for the offline stage, we report the total token usage per dataset in Table~\ref{tab:llm_tokens}. Since absolute dollar estimates may vary with changes in API pricing, we use token counts and relative reductions as more stable measures of cost efficiency.

For online inference, we compare RAG-HAR+ with RAG-HAR in terms of token usage. In RAG-HAR, every test sample is sent to the LLM after retrieval, so the online cost grows directly with the number of test samples. In contrast, RAG-HAR+ invokes the Ambiguity Resolver Agent only when majority voting over the retrieved neighbors is undecided. To make the comparison clear, we report both the total token usage over the full test set and the average token usage per test sample. Total token usage reflects the actual cost of evaluating each dataset, while per-sample token usage normalizes for differences in test-set size. These results are summarized in Table~\ref{tab:online_llm_tokens} and Fig.~\ref{fig:per-sample-cost}(a). The per-sample prompt size generally increases with the number of sensor channels, since more channel-wise descriptors must be included in the prompt. For example, the Skoda dataset has 60 channels, resulting in approximately 132.9K tokens (per sample), which fits comfortably within the 400k-token context window of  \texttt{gpt-5-mini}~\cite{openai_gpt5mini_2026}.

\begin{table}[b]
\centering
\caption{Offline LLM token usage by Retrieval Designer Agent in RAG-HAR+. This one-time cost is incurred once per dataset configuration.}
\label{tab:llm_tokens}
\footnotesize
\setlength{\tabcolsep}{4pt}
\renewcommand{\arraystretch}{1.05}
\begin{tabular}{lrrr}
\toprule
\textbf{Dataset} & \textbf{Input} & \textbf{Output} & \textbf{Total} \\
\midrule
USC-HAD & 17,929 & 3,996 & 21,925 \\
PAMAP2  & 18,011 & 3,536 & 21,547 \\
MHEALTH & 18,985 & 4,044 & 23,029 \\
GOTOV   & 18,546 & 5,095 & 23,641 \\
HHAR    & 16,712 & 3,427 & 20,139 \\
Skoda   & 18,084 & 3,811 & 21,895 \\
\midrule
\textbf{Total} & \textbf{108,267} & \textbf{23,909} & \textbf{132,176} \\
\bottomrule
\end{tabular}
\vspace{-4mm}
\end{table}

\begin{table*}[t]
\centering
\caption{Online LLM token usage comparison between RAG-HAR and RAG-HAR+ for each dataset. In. = input tokens, Out. = output tokens, Total = total tokens, and LLM Calls = number of inference-time LLM invocations. Reduction is computed from total tokens. } 
\label{tab:online_llm_tokens}
\scriptsize
\setlength{\tabcolsep}{5.5pt}
\renewcommand{\arraystretch}{1.05}
\begin{tabular}{lrrrrrrrrrr}
\toprule
\multirow{2}{*}{\textbf{Dataset}}
& \multirow{2}{*}{\textbf{Test}}
& \multicolumn{4}{c}{\textbf{RAG-HAR}}
& \multicolumn{4}{c}{\textbf{RAG-HAR+}}
& \multirow{2}{*}{\shortstack{\textbf{Token Usage}\\\textbf{Reduction (\%)}}} \\
\cmidrule(lr){3-6} \cmidrule(lr){7-10}
&
& \textbf{In.}
& \textbf{Out.}
& \textbf{Total}
& \textbf{LLM Calls}
& \textbf{In.}
& \textbf{Out.}
& \textbf{Total}
& \textbf{LLM Calls}
& \\
\midrule
USC-HAD  & 492  &  6,734,004  & 617,066   & 7,351,070  & 492  & 715,790 & 68,667 & 784,457 & 53 & 89.3 \\
PAMAP2 & 342 & 22,820,018 & 348,395 & 23,168,414 & 342 & 333,414 & 4,642 & 338,056 & 5 & 98.6 \\
MHEALTH  & 666  & 34,587,378 & 781,484 & 35,368,862 & 666  & 52,250     & 597    & 52,847    & 1 & 99.9 \\
GOTOV    & 1324 & 34,883,029 & 1,101,832 & 35,984,861 & 1324 & 1,400,776 & 89,228 & 1,490,004 & 53 & 95.9 \\
HHAR     & 858  & 12,861,077  & 637,237 & 13,498,314  & 858  & 641,636   & 73,329   & 714,965   & 65 & 94.7 \\
Skoda    & 793  & 104,775,918 & 639,317 & 105,415,235 & 793  & 1,639,389         & 16,422     & 1,655,811         & 14 & 98.4 \\
\midrule
Total & 4475 & 216,661,424 & 4,125,331 & 220,786,756 & 4475
& 4,783,255 & 252,885 & 5,036,140 & 191 & 97.7 \\
\bottomrule
\end{tabular}
\vspace{-6mm}
\end{table*}

The offline feature-group design stage uses 132{,}176 tokens across all six datasets, corresponding to an estimated total cost of \$0.075. Since this cost is incurred only once per dataset configuration, it is small compared with the recurring online inference cost. For online inference, RAG-HAR uses 220.8M tokens across the six test sets. RAG-HAR+ reduces this to 5M tokens by routing only ambiguous samples to the Ambiguity Resolver Agent. This corresponds to a 97.7\% reduction in online LLM token usage. The reduction varies across datasets because the fallback rate depends on label agreement among the retrieved neighbors. When the selected feature groups produce consistent retrieved neighbors, most samples are classified directly through retrieval-based voting. MHEALTH represents the strongest case, where only one sample is routed to the Ambiguity Resolver Agent, resulting in only one online LLM classification cost. Even in USC-HAD, which shows the smallest reduction, RAG-HAR+ still reduces online cost by 89.3\%. These results show that RAG-HAR+ substantially reduces recurring LLM usage while preserving the retrieval-augmented classification framework.


Fig.~\ref{fig:accuracy_contribution} further separates the contribution of retrieval-based classification from that of the Ambiguity Resolver Agent. Across all datasets, most of the final classification F1 is achieved directly through retrieval-based voting. Compared with RAG-HAR, RAG-HAR+ improves the retrieval-only F1 and narrows the gap between retrieval-based prediction and final LLM-assisted F1, indicating that the adaptive feature-group retriever provides stronger evidence before fallback. The Ambiguity Resolver Agent then adds 0.1--3.3 percentage points, with the largest gains on USC-HAD and HHAR. These results show that RAG-HAR+ resolves most predictions without invoking the Ambiguity Resolver Agent, while retaining it as a useful fallback for ambiguous samples. Thus, selective fallback substantially reduces online LLM usage while preserving the accuracy gains provided by LLM-based reasoning.
RAG-HAR+ does not eliminate LLM usage entirely; instead, it changes where and how the LLM is used. A small offline LLM cost is invested in constructing a stronger retrieval representation, while online LLM calls are reserved for cases where retrieval evidence is insufficient. This makes the inference cost scale with ambiguity rather than with the total number of test samples. The reported costs account only for LLM API usage. They do not include non-LLM components such as preprocessing, feature extraction, vector indexing, nearest-neighbor retrieval, storage, or system overhead.

\subsection{Inference latency}
\label{subsec:latency}

Fig.~\ref{fig:per-sample-cost}(b) compares the per-sample latency of RAG-HAR and RAG-HAR+. RAG-HAR+ reduces latency on every dataset, with speed-ups ranging from $5.7\times$ on HHAR to $44.5\times$ on MHEALTH. This reduction comes from the different inference strategy used by RAG-HAR+. In the RAG-HAR pipeline, every test window is routed to the LLM after retrieval, which introduces a network round-trip and LLM generation delay for each sample. In contrast, RAG-HAR+ first attempts retrieval-based classification and routes only ambiguous samples to the Ambiguity Resolver Agent. As a result, most samples avoid the online LLM stage, and the per-sample latency moves closer to the cost of feature computation and retrieval alone.

The speed-up varies across datasets because the cost of LLM-based inference and retrieval-based inference differs with dataset characteristics. MHEALTH obtains the
largest reduction, from 18.25\,s to 0.41\,s per sample
($44.5\times$), because only one test sample is routed to the
Ambiguity Resolver Agent. Skoda achieves the second-largest
speed-up of $22.1\times$, while HHAR shows the smallest, but
still substantial, speed-up of $5.7\times$. Nevertheless, RAG-HAR+ still reduces latency substantially on this dataset.
The remaining latency of RAG-HAR+ mainly reflects the retrieval pipeline, including feature extraction over the query window, feature normalization, and approximate nearest-neighbor search over the indexed training corpus. This cost depends jointly on the number of channels, window length, selected feature groups, and index size, rather than on channel count alone. For example, USC-HAD shows the highest RAG-HAR+ latency, while Skoda remains near the lower end despite having the largest number of channels. Across all datasets, the cost-optimized pipeline keeps latency between 0.41s and 2.04s per sample, which is compatible with continuous, near-real-time activity recognition on mobile and edge devices.

\begin{figure}
    \centering
    \includegraphics[width=\linewidth]{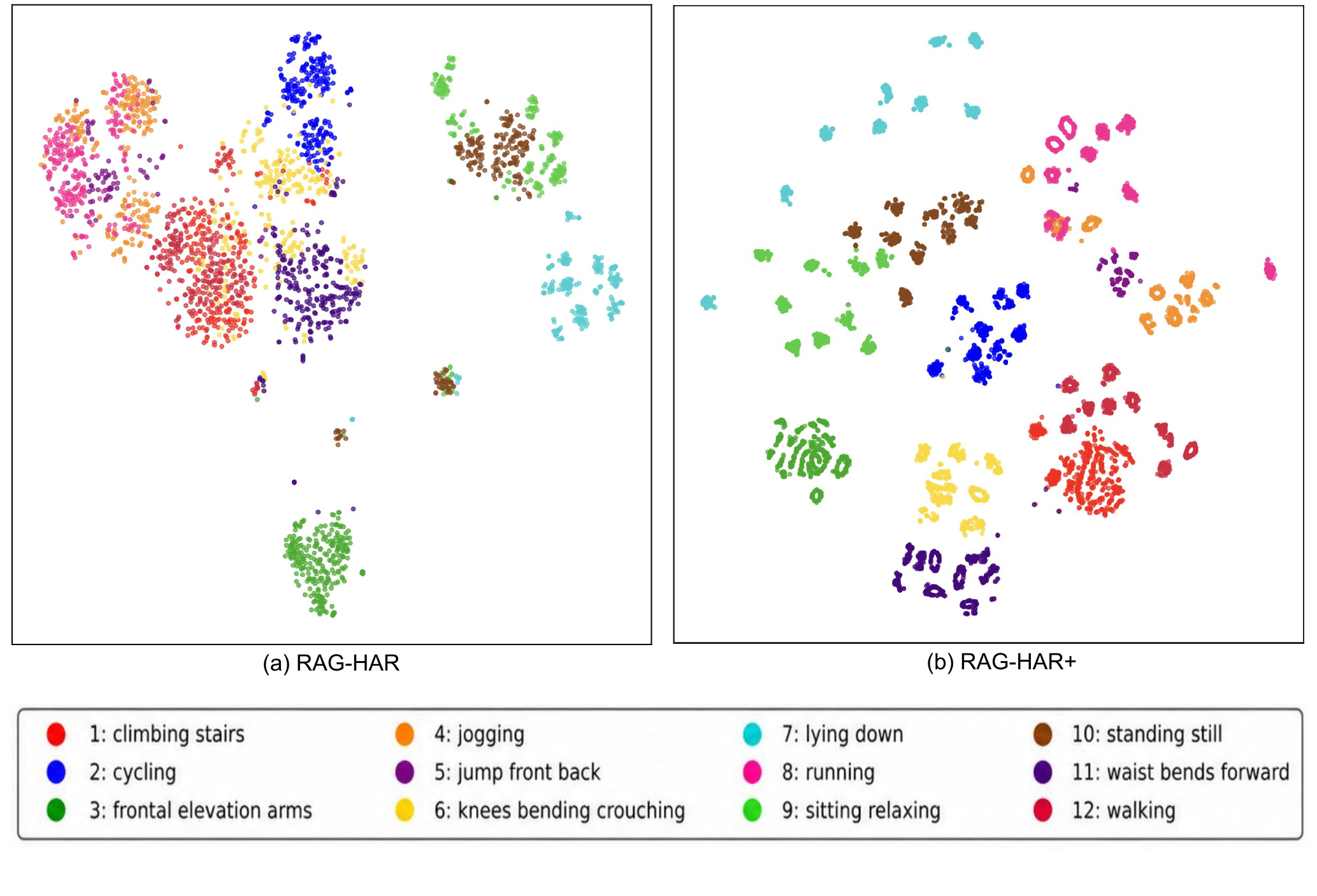}
    \vspace{-8mm}
    \caption{t-SNE analysis of RAG-HAR vs. RAG-HAR+ retriever performance on MHEALTH.}
    \label{fig:retriever_tsne}
    \vspace{-4mm}
\end{figure}

\subsection{Retriever Impact: RAG-HAR+ vs. RAG-HAR}
\label{sec:retriever_performance}

This subsection analyzes how the redesigned RAG-HAR+ retriever improves retrieval quality over the original RAG-HAR retriever. RAG-HAR represents each window using fixed temporal segment views with the same basic statistical descriptors, whereas RAG-HAR+ uses the Retrieval Designer Agent to select three dataset-specific feature groups from statistical, temporal, spectral, and signal-shape descriptors. These groups form complementary vector fields in the multi-vector index, allowing retrieval to capture activity-relevant cues that fixed temporal summaries may miss.

Fig.~\ref{fig:accuracy_contribution} compares the RAG-HAR retriever with the final optimized RAG-HAR+ retriever produced by the Retrieval Designer Agent. For RAG-HAR, RF1 is reported only as a diagnostic retrieval-quality measure computed by applying voting over its retrieved neighbors; the original RAG-HAR pipeline still invokes the LLM for final classification of every sample. In contrast, RAG-HAR+ RF1 reflects its actual retrieval-based voting path before fallback. Across datasets, RAG-HAR+ improves RF1 and maintains high RHR, indicating that the correct activity label is usually present among the retrieved neighbors. The reduced gap between RF1 and final LLM-assisted F1 further shows that the adaptive retriever provides stronger evidence before fallback, allowing most samples to be classified without invoking the Ambiguity Resolver Agent.

Fig.~\ref{fig:retriever_tsne} provides a qualitative view of this effect using MHEALTH retrieval embeddings. The RAG-HAR retriever produces a more entangled embedding space, where several activity classes overlap because fixed segment-level statistics capture only coarse window-level patterns. In contrast, the RAG-HAR+ retriever forms more coherent and better separated clusters, showing that the selected feature groups capture activity-specific retrieval cues more effectively. This observation is consistent with Fig.~\ref{fig:accuracy_contribution}: on MHEALTH, RAG-HAR+ improves retrieval F1 from $82.9\%$ to $98.1\%$ and increases RHR from $95.7\%$ to $99.2\%$. A more separable retrieval space increases the likelihood that the Top-$q$ neighbors agree on the correct label, enabling direct classification through voting.

The t-SNE visualization also explains why RAG-HAR+ reduces online LLM usage. When clusters are compact and well separated, the deterministic routing rule can identify confident retrieval evidence and avoid the latency and cost of an LLM call. However, Fig.~\ref{fig:retriever_tsne}(b) also shows that some dynamic activities with similar repetitive motion patterns remain close in the embedding space. These boundary cases justify retaining the Ambiguity Resolver Agent. Thus, RAG-HAR+ does not remove the LLM entirely; it uses retrieval for clear cases and reserves the LLM for genuinely ambiguous ones.

\subsection{Ablation Studies}

\begin{figure*}[t]
    \centering
    \includegraphics[width=1\linewidth]{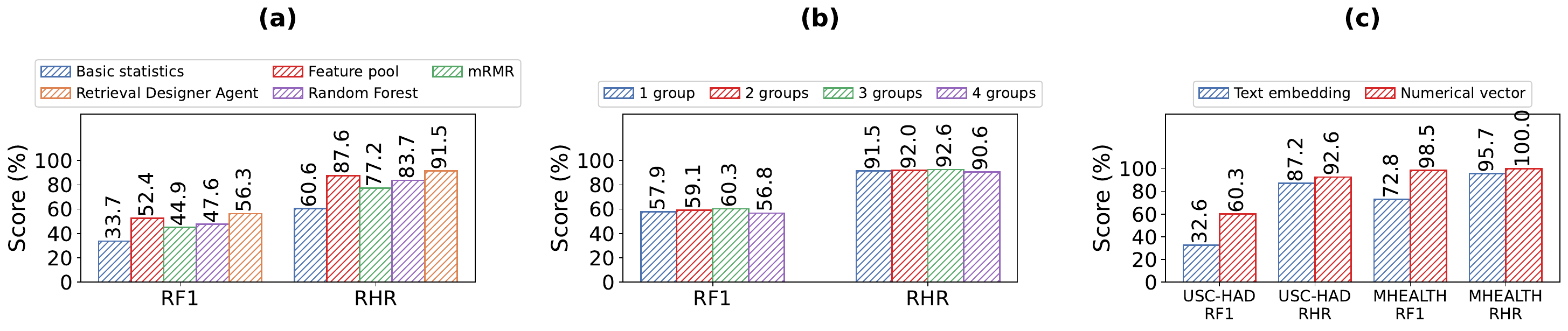}
    \vspace{-6mm}
    \caption{Ablation of the offline retrieval design.
(a) Held-out USC-HAD test performance of the basic-statistics,
all-candidate, mRMR-selected, and Retrieval Designer Agent-selected
feature sets.
(b) USC-HAD validation performance for different feature groups.
(c) Validation performance of vectorizer
index representations on USC-HAD and MHEALTH. }
    \label{fig:ab-results-set}
    \vspace{-4mm}
\end{figure*}

This subsection evaluates the principal design choices of RAG-HAR+. We first evaluate the contribution of the adaptive feature-group design. We then evaluate the
representation used to construct the retrieval index and analyze the
combined effectiveness of the resulting retriever. Finally, we study
the online inference stage by varying the number of retrieved neighbors
and the textual representation used by the Ambiguity Resolver Agent.

\subsubsection{Offline Retrieval Designer Agent}

\paragraph{\textbf{Feature-Group Selection}}
\label{sec:ablation_features_for_data_representation}

We evaluate whether the Retrieval Designer Agent selects genuinely useful retrieval features, or whether the gains come only from using a richer candidate pool than the basic statistical representation. To isolate feature selection from grouping effects, all configurations in this experiment are concatenated into a single numerical vector and evaluated under the same train/validation/test splits, normalization, similarity function, and retrieval depth. 
On USC-HAD, the Retrieval Designer Agent selects 31 features in total; therefore, the budget-matched baselines also select 31 features from the same 67-feature candidate pool. Fig.~\ref{fig:ab-results-set}(a) reports the results. Three observations follow. First, expanding the feature pool improves retrieval quality: using all 67 candidate features increases RF1 from 33.7\% to 52.4\%, showing that richer motion descriptors provide more discriminative retrieval cues than basic statistics alone. Second, feature selection further improves performance beyond pool expansion. The agent-selected 31 features achieve 56.3\% RF1, outperforming the full 67-feature pool while retaining less than half of the candidate features. This suggests that the agent removes noisy or redundant descriptors rather than simply benefiting from a larger representation. Third, the Retrieval Designer Agent outperforms both classical supervised filter-based selection and trained model-based selection: minimum Redundancy Maximum Relevance (mRMR) achieves 44.9\% RF1, while Random-Forest-based feature selection achieves 47.6\%. Thus, the LLM-guided retrieval design provides a stronger training-free alternative to conventional feature selection, improving retrieval quality without training a dataset-specific classifier.

\paragraph{\textbf{Number of Feature Groups}}

\label{sec:ablation_number_of_feature_views}
We analyze the effect of the number of feature groups used for retrieval in RAG-HAR+. Since different activity patterns may be better captured by different feature groups, using multiple feature groups can improve the diversity and relevance of retrieved examples. However, increasing the number of feature groups may also introduce redundant or less informative retrieval evidence.  Therefore, we conduct an ablation study on the USC-HAD dataset by varying the number of feature groups from 1 to 4 while keeping all other components unchanged.

Fig. \ref{fig:ab-results-set}(b) reports the validation retrieval
F1 score and RAG hit rate for each configuration. The validation retrieval F1 score increases from $K{=}1$ to $K{=}3$, reaching its highest value at $K{=}3$, and then drops at $K{=}4$. This suggests that
three feature groups provide sufficient complementary information without introducing excessive redundancy or fragmenting the per-group feature pool. Based
on this result, we use $K{=}3$ for all subsequent experiments.

\paragraph{\textbf{Feature-Group Assignment}}
\label{sec:ablation_grouping}

The agent decides not only \emph{which} features to select but also
\emph{how} to partition them into the three vector fields. To isolate the
contribution of this assignment, we fix the exact 31 features selected for
USC-HAD and vary only their organization, evaluating all conditions with
equal group weights so that weight optimization cannot mask grouping
effects. The agent's grouping achieves a test retrieval F1 of 56.1\%,
compared with 54.6\,$\pm$\,1.6\% for 20 random partitions into the same
group sizes and 55.3\% for a single concatenated vector. We conclude that the
benefit of the Retrieval Designer Agent is attributable to the specific group assignment as well as feature selection.

\paragraph{\textbf{Feature-Group Weighting}}
\label{sec:ablation_feature_group_weights}
We next evaluate the effect of feature-group weighting in Table~\ref{tab:weight_search}. The number of feature groups is fixed at three, while their relative contributions to the retrieval score are varied. Because the selected feature groups capture different characteristics of human motion, they
may not contribute equally to retrieving relevant samples. We therefore select dataset-specific weights rather than assigning equal importance
to all three groups. The weights are selected through a grid search performed on a validation set as part of the Retrieval Designer Agent. The validation set is constructed exclusively from the training split, ensuring that the test split remains unseen during weight selection. For the multi-subject datasets MHEALTH, HHAR, PAMAP2, GOTOV and USC-HAD, two complete subjects are held out from the training split for validation. 
For Skoda, we use a random 20\% window-level validation split with a fixed seed.

For each dataset, the weights $(w_1,w_2,w_3)$ are constrained to sum to one, and each weight is varied from 0 to 1 in increments of 0.2. The
configuration that achieves the highest validation retrieval F1-score is selected and fixed for the final test-set evaluation. This procedure is performed independently for each dataset.

\begin{table}[t]
\centering
\caption{Validation performance for different feature-group weight combinations on the USC-HAD validation set.}
\vspace{-2mm}
\label{tab:weight_search}
\begin{tabular}{ccccc}
\toprule
$w_1$ & $w_2$ & $w_3$ & Val RF1 & Val RHR \\
\midrule
\textbf{0.2} & \textbf{0.6} & \textbf{0.2} & \textbf{60.33} & \textbf{92.60} \\
0.4 & 0.4 & 0.2 & 60.12 & 92.60 \\
0.6 & 0.2 & 0.2 & 58.83 & 89.81 \\
0.2 & 0.4 & 0.4 & 58.13 & 90.84 \\
0.2 & 0.2 & 0.6 & 57.34 & 89.40 \\
0.4 & 0.2 & 0.4 & 56.04 & 88.17 \\
\bottomrule
\end{tabular}
\vspace{-4mm}
\end{table}

\subsubsection{Offline Index Construction}

\paragraph{\textbf{Index Representation}}
\label{sec:ablation_vectorizer_for_data_representation}

We analyze whether retrieved samples should be represented as text embeddings or numerical vectors. Fig.~\ref{fig:ab-results-set}(c) evaluates the effect of vector representation. To isolate the effect of the retrieval representation, both settings use the same LLM-selected feature groups. We then compare two ways of storing and retrieving the resulting feature values: representing them as textual descriptions and embedding them using a text embedding model versus directly storing the normalized feature values as numerical vectors.
 This isolates whether numerical vector retrieval is more suitable than text-embedding retrieval for sensor-feature matching. The results show that the retriever benefits more from numerical vectors than from language-style text embeddings.

\subsubsection{Ambiguity Resolver Agent}
\paragraph{\textbf{Number of Retrieved Neighbors}}
\label{sec:ablation_optimal_k}

\begin{figure}[t]
    \centering
    \includegraphics[width=1\linewidth]{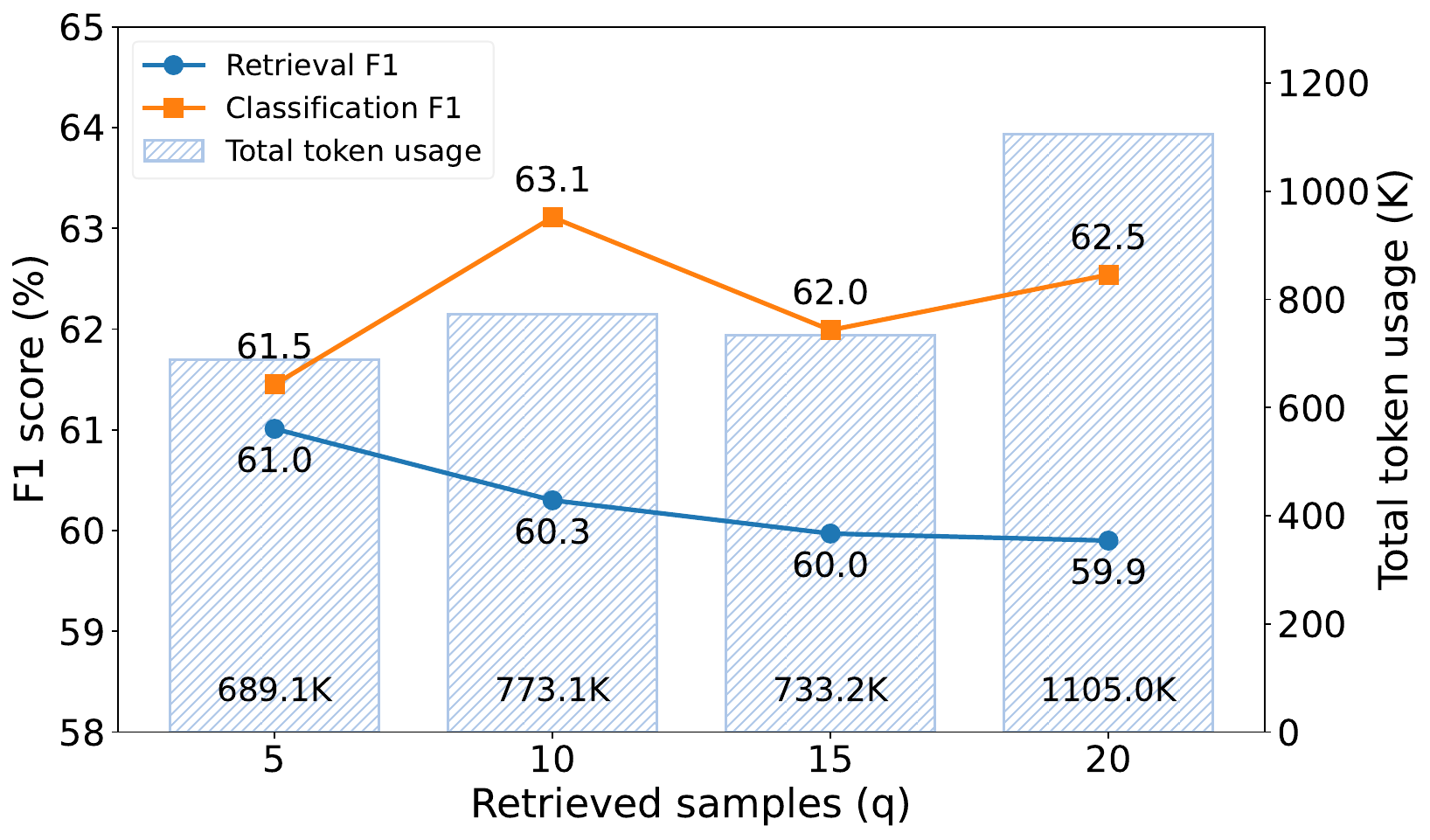}
    \vspace{-6mm}
    \caption{Effect of retrieval depth $q$ on USC-HAD validation performance and token usage.}
    \label{fig:k-ab}
    \vspace{-4mm}
\end{figure}

We analyze the effect of the number of retrieved samples, $q$, on the performance and efficiency of RAG-HAR+. Since retrieved samples are used both for retrieval-based classification and as contextual examples in the Ambiguity Resolver Agent prompt, increasing $q$ can provide richer evidence but also increases prompt length and may introduce less relevant neighbors. As shown in Fig.~\ref{fig:k-ab}, $q=10$ achieves the best final classification F1-score of 63.1\%, improving over $q=5$ while maintaining similar retrieval quality. Larger values of $q$ do not provide further gains. Although $q=15$ slightly reduces total token usage compared with $q=10$ (733.2K vs. 773.1K tokens), this saving is relatively small and comes with a drop in final F1-score from 63.1\% to 62.0\%. At $q=20$, the average tokens per LLM call increase further, resulting in the highest total token usage among the evaluated settings. Overall, $q=10$ provides the best accuracy-efficiency trade-off, achieving the highest final classification F1-score while keeping total token usage within a reasonable range. Therefore, we use 10 as the retrieval depth in RAG-HAR+.

\begin{figure}[t]
    \centering
    \includegraphics[width=1\linewidth]{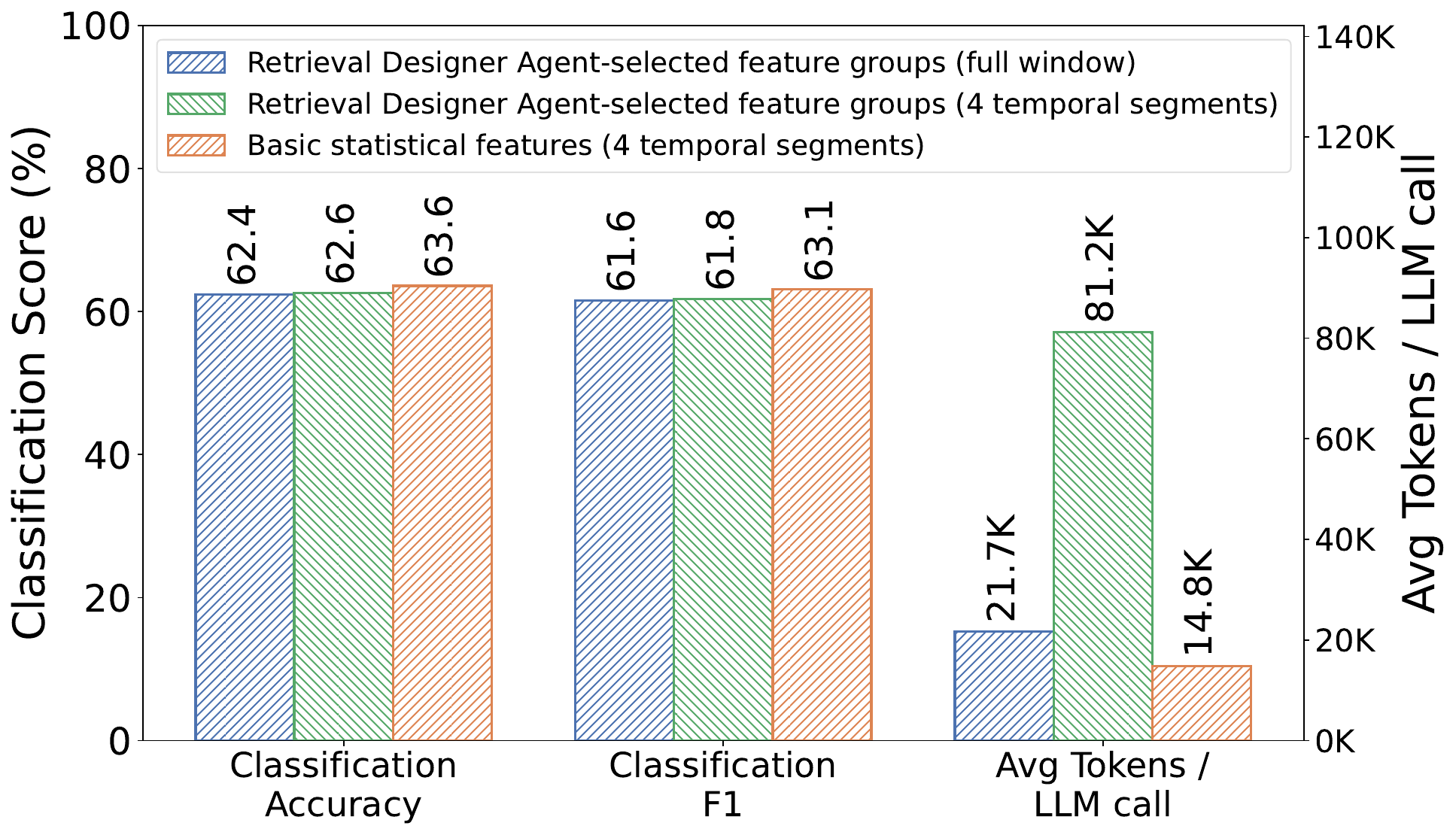}
    \vspace{-6mm}
    \caption{USC-HAD validation-set ablation of the textual representation supplied to the Ambiguity Resolver Agent. }
    \label{fig:llm-prompt-ab}
    \vspace{-4mm}
\end{figure}

\paragraph{\textbf{Ambiguity Resolver Agent Prompt Representation}}
In the Ambiguity Resolver Agent, the LLM receives the test sample to be classified together with the retrieved samples from the vector database. Since raw signals are not directly suitable for LLM input, both the test sample and the retrieved samples are converted into textual feature descriptions before being included in the prompt. This ablation studies how different prompt-level feature representations affect the classifier's performance and token usage. We compare three feature representation strategies. First, in the \textit{selected feature-group representation}, the feature groups selected by the Retrieval Designer Agent are computed over the full sensor window and represented as feature descriptions in the form \texttt{feature = value}. Second, in the \textit{segment-wise selected feature-group representation}, the same selected feature groups are computed separately over four temporal segments: start, middle, end, and full window. These segment-level values are then arranged as textual feature descriptions. Third, in the \textit{segment-wise basic statistical representation}, basic statistical features are computed over the start, middle, end, and full-window segments, following the representation of \textit{feature = value}. As shown in Fig.~\ref{fig:llm-prompt-ab}, the Retrieval Designer Agent-selected feature-group representation achieves 62.4\% accuracy and 61.6\% F1-score, with an average token usage of 21.7K tokens per LLM call. Extending the selected feature groups across temporal segments slightly improves performance to 62.6\% accuracy and 61.8\% F1-score, but substantially increases the average token usage to 81.2K tokens per LLM call. In contrast, the segment-wise basic statistical representation achieves the best classification performance, with 63.6\% accuracy and 63.1\% F1-score, while requiring the lowest token usage of 14.8K tokens per LLM call. This result supports separating the retrieval representation from the LLM prompt representation: adaptive feature groups improve neighbor retrieval, while compact segment-wise statistical summaries provide a more effective and token-efficient textual representation for the Ambiguity Resolver Agent.

\subsection{Discussion}\label{subsec:discussion}

The effectiveness of RAG-HAR+ is primarily governed by retrieval quality. RAG-HAR+ deliberately uses a simple deterministic majority-voting rule for routing. This choice keeps the inference path parameter-free, reproducible, and training-free, which is important for deployment across new datasets without requiring calibration data. However, the same design also opens a useful direction for future work. Rather than routing only when the top retrieved labels tie, future systems could use a more adaptive uncertainty-estimation strategy based on the full retrieval distribution. For example, the entropy of the neighbour-label distribution, the vote margin between the top two classes, the agreement across feature groups, or the similarity-score concentration could be combined to decide when LLM fallback is worthwhile. Such an entropy-based LLM fallback strategy could better balance accuracy and cost by invoking the LLM for genuinely uncertain cases while avoiding unnecessary calls when retrieval evidence is already strong. As future work, we will investigate the privacy implications of sharing sensor data with cloud-hosted LLMs; in this version, we scope the study to computational cost and latency reduction.

\section{Pilot Mobile Prototype Implementation}
\label{sec:pilot_mobile_prototype}

\begin{figure*}[t]
    \centering
    \includegraphics[width=1\linewidth]{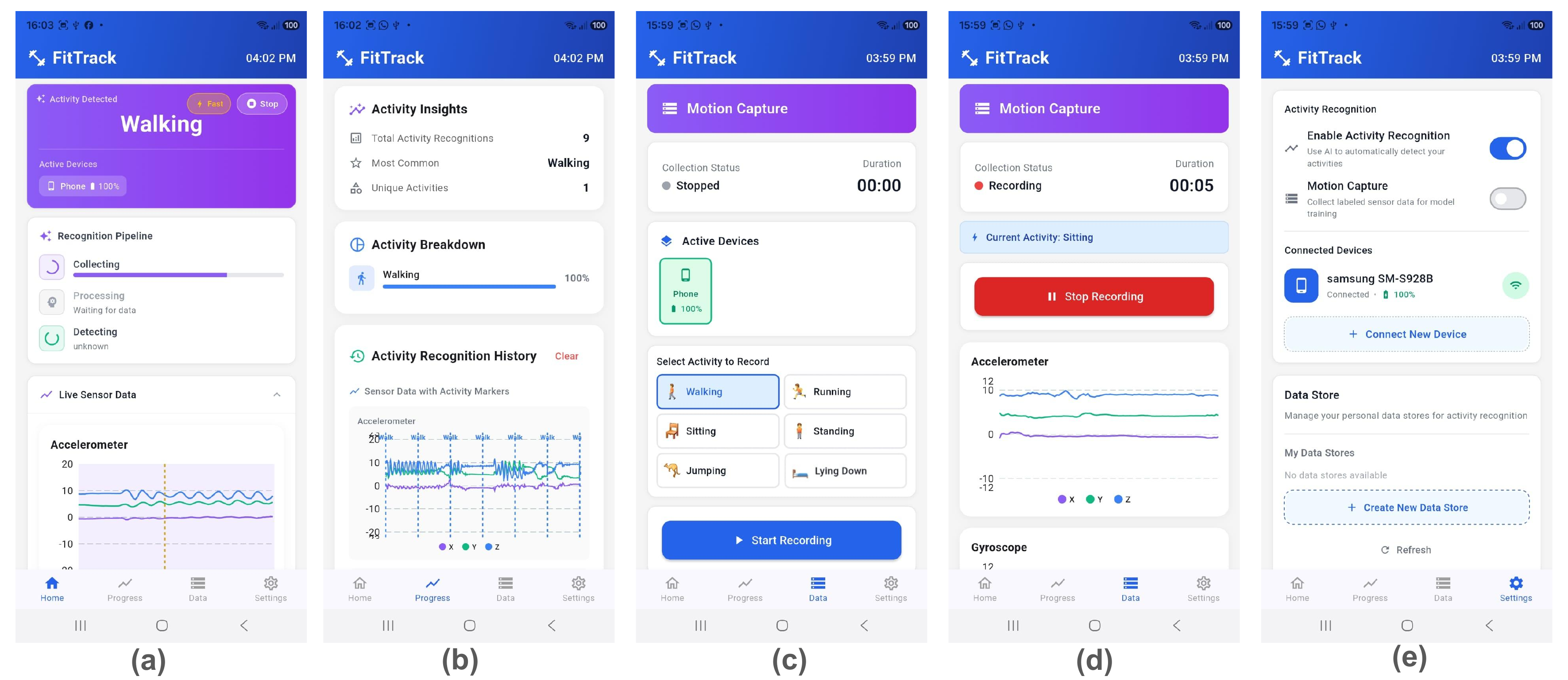}
    \vspace{-6mm}
    \caption{RAG-HAR+ widget embedded within a mobile fitness-tracking application: (a) activity prediction, (b) activity insights and history, (c) labeled motion-data collection, (d) real-time motion capture, and (e) application settings and device/data-store management.}
    \label{fig:demo-row}
    \vspace{-4mm}
\end{figure*}

To examine the practical deployability of RAG-HAR+, we implement a pilot mobile prototype~\cite{11585308} as an activity-recognition widget within a Flutter-based fitness-tracking application~\cite{flutter}. The prototype runs on a Samsung Galaxy S23 Ultra smartphone equipped with a Snapdragon 8 Gen 2 Mobile Platform for Galaxy, 12\,GB RAM, and built-in accelerometer and gyroscope sensors. The prototype separates offline preparation from online mobile inference. The laptop-hosted backend runs the RAG-HAR+ Retrieval Designer Agent once on mobile-collected labeled data to design feature groups and build the multi-vector Milvus index~\cite{milvus}. At deployment, the smartphone runs the RAG-HAR+ inference pipeline in real time, invoking the remotely hosted Ambiguity Resolver Agent only for rare ambiguous cases.

The application also supports labeled data collection and incremental retrieval-index expansion. As shown in Fig.~\ref{fig:demo-row}, users can record labeled motion samples through the mobile interface, after which the corresponding feature vectors and labels can be added to the retrieval index without retraining a classifier. The Retrieval Designer Agent only needs to be rerun when the sensing configuration, activity set, or candidate feature pool changes substantially. Thus, the prototype demonstrates how RAG-HAR+ can support mobile activity recognition while avoiding continuous transmission of raw sensor streams and reducing unnecessary online LLM calls.

\begin{figure}[t]
    \centering
    \includegraphics[width=1\linewidth]{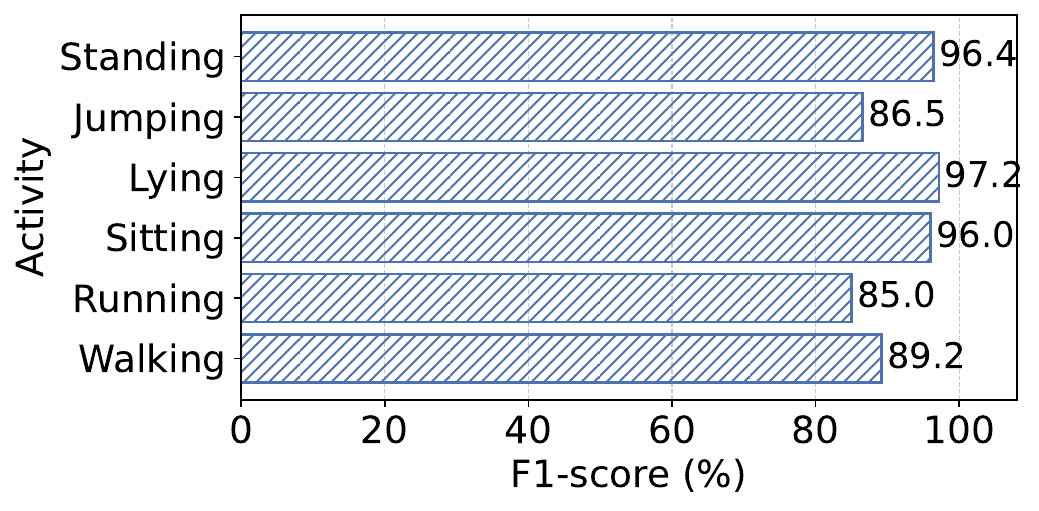}
    \vspace{-6mm}
    \caption{Per-activity F1-scores of the mobile-integrated RAG-HAR+ prototype on smartphone-collected activity data.}
    \label{fig:mobile-eval}
    \vspace{-4mm}
\end{figure}

We evaluate the prototype on smartphone-collected activity data to assess runtime behavior under real deployment rather than benchmark conditions. RAG-HAR+ slightly improves macro-F1 compared with RAG-HAR ($91.7\%$ vs.~$91.0\%$) while reducing average per-sample latency from $5400$ to $360$\,ms, corresponding to a roughly $15\times$ speed-up. This latency reduction is larger than in the benchmark experiments because the prototype captures real network round-trips to the hosted LLM: RAG-HAR incurs this cost for every prediction, whereas RAG-HAR+ resolves most windows locally through neighbor voting and routes only ambiguous cases to the Ambiguity Resolver Agent. Fig.~\ref{fig:mobile-eval} shows stronger recognition for static and postural activities and lower performance for dynamic locomotion activities. 
Overall, the pilot prototype maintains benchmark-level accuracy while substantially reducing per-sample latency, supporting the feasibility of retrieval-first, LLM-assisted HAR in a real mobile setting.

\section{Conclusion}\label{sec:conclusion}

RAG-HAR introduced a training-free HAR framework that combines retrieval-augmented generation with LLM reasoning, enabling activity recognition through similarity search over sensor-window descriptors and contextual reasoning over retrieved examples, without dataset-specific training or fine-tuning. This direction is timely beyond HAR as well, since recent surveys in adjacent domains show rapidly growing research activity around LLM-based methods \cite{nirhoshan2026survey}. Compared with prior supervised HAR pipelines, RAG-HAR achieved state-of-the-art performance across six benchmarks and further demonstrated open-set reasoning by inferring unseen activities using the LLM's prior knowledge. RAG-HAR+ extends this paradigm toward practical edge deployment by strengthening retrieval and changing the role of the LLM from a per-sample online classifier to an offline Retrieval Designer Agent and an online Ambiguity Resolver Agent used only for difficult cases. Across six benchmarks, RAG-HAR+ preserves competitive or improved recognition performance while reducing online LLM usage by $89.3$--$99.9\%$ and lowering per-sample latency by up to an order of magnitude, with a smartphone prototype confirming the feasibility of retrieval-first, LLM-assisted HAR in mobile settings. Overall, RAG-HAR and RAG-HAR+ show that retrieval, LLM reasoning, and cost-aware routing can provide a scalable, training-free alternative to conventional HAR pipelines, where inference cost scales with ambiguity rather than data volume. 

\bibliographystyle{IEEEtran}
\bibliography{refs}
\end{document}